# Symbolic Decision Theory and Autonomous Systems

John Fox and Paul Krause
Biomedical Computing Unit
Imperial Cancer Research Fund Laboratories
Lincoln's Inn Fields, London, United Kingdom

## Abstract

The ability to reason under uncertainty and with incomplete information is a fundamental requirement of decision support technology. In this paper we argue that the concentration on theoretical techniques for the evaluation and selection of decision options has distracted attention from many of the wider issues in decision making. Although numerical methods of reasoning under uncertainty have strong theoretical foundations, they are representationally weak and only deal with a small part of the decision process. Knowledge-based systems, on the other hand, offer greater flexibility but have not been accompanied by a clear decision theory. We describe here work which is under way towards providing a theoretical framework for *symbolic decision procedures.* A central proposal is an extended form of inference which we call *argumentation*; reasoning for and against decision options from generalised domain theories. The approach has been successfully used in several decision support applications, but it is argued that a comprehensive decision theory must cover autonomous decision making, where the agent can formulate questions as well as take decisions. A major theoretical challenge for this theory is to capture the idea of *reflection* to permit decision agents to reason about their goals, what they believe and why, and what they need to know or do in order to achieve their goals.

## 1 INTRODUCTION[1]

Medicine has been an important field for developing and testing decision support systems which are capable of reasoning with uncertain and incomplete information. Experiments with numerical techniques for diagnosis and other applications began in the sixties, and produced some early encouraging progress, notably de Dombal's classic work on the diagnosis of abdominal pain [1]. By the end of the seventies, systems such as MYCIN, INTERNIST and CASNET were showing that symbolic techniques for knowledge representation, inference and heuristic reasoning held much promise for decision support systems. While arousing great excitement, these early expert systems were also treated with some skepticism by decision theorists on the grounds that they were somewhat ad hoc in design. This stimulated a great deal of technical activity in developing more rigorous and precise numerical uncertainty handling techniques, but we feel that this has distracted attention from many fundamental issues which still need to be addressed in order to produce flexible and sound decision support systems that will have significant impact in many practical applications.

A major cause of the criticisms levelled at some of the above mentioned systems, was the attempt to incorporate uncertainty handling into a simple rule-based knowledge representation framework. There is, unfortunately, a fundamental conflict between the demands of computational tractability and of semantic expressiveness. The modularity of simple rule-based systems aids efficient data update procedures. However, severe evidence independence assumptions have to be made for uncertainties to be combined and propagated using strictly local calculations. A general and rigorous implementation of a fully intentional system, in which all possible interactions between rules and evidences are taken into account on each data update, could become so computationally intractable that the development of a realistic application would be infeasible

---

1. This paper is a shortened and revised version of a keynote address given at the IMACS workshop in Qualitative Reasoning and Decision Support Systems, Toulouse, March 13-15, 1991 [8].

[12]. In order to develop computationally tractable, yet rigorous, uncertainty handling mechanisms much recent work has been directed towards the development of graphical structures in which the dependencies and influences between knowledge items are explicitly represented [11], [13]. This has led to a realisation that the correct structuring of the knowledge that is relevant in a given decision making context is as important, *if not more important,* than the numerical values that are propagated through the graph.

In fact, at least for certain classes of applications such as medical diagnostic applications, decision accuracy can be highly insensitive to these values. In [2] for example the performance of a strictly probabilistic approach to diagnosis and a heuristic approach were quantitatively compared. Using a database of medical records of some 400 patients who had been reliably diagnosed as having one of 5 different gastrointestinal conditions, two diagnostic systems were constructed. The first was a simple bayesian procedure for computing posterior probabilities given a set of patient symptoms. The second was a set of categorical production rules for interpreting patterns of symptoms. The rules excluded quantitative information about the associations between symptoms and diseases. A typical rule was:

    if:    age(elderly) and weight_loss(present)
    then:  maybe(cancer)

It turned out that the diagnostic accuracy of the rule-based system approximately equalled that of the probabilistic system (~70%)[1] while requesting only half the available symptom data. The relative naturalness of the categorical representation did not apparently entail a significant reduction in decision making performance.

There are well documented differences between the performance of different numerical calculi [17], [18]. However, theoreticians have not paid so much attention to studies comparing precise with imprecise methods like the above. There is considerable evidence that the performance of a well structured, largely symbolic system may well be as good as a more rigorous numerical approach [20]. In [28], it was demonstrated that a purely symbolic system only differed in behaviour from a numerical probabilistic system in those cases with an uncommon diagnosis (prior probability ≤ 0.03). Chard's conclusion was that, so long as it could be ensured that a purely symbolic approach could pick up the less common conditions, then it would suffer no performance disadvantage when compared to a bayesian system.

We will go further than simply saying there is no disadvantage; we shall argue that a symbolic approach to decision making has in fact many advantages (other than reducing the purely computational and cost overheads associated with the elicitation and use of large amounts of numerical data). The next section will suggest that a symbolic approach allows us to explicate more of the decision process, including the knowledge required to define, organise and make a decision. It also allows us to explicitly represent decisions, knowledge sources, reasoning strategies and representations, and to reason about the control and inference processes involved in specific tasks. These are all requirements which need to be satisfied if we are to be able to develop systems with an advanced decision making capability. The third section of this paper addresses the requirements of a particularly challenging class of AI system, those capable of making decisions autonomously.

## 2 SYMBOLIC DECISION MAKING UNDER UNCERTAINTY

The *Concise Oxford Dictionary* defines a decision as follows "Decision: settlement of (question etc.), conclusion, formal judgement, making up one's mind". But how should we settle questions, particularly where they involve uncertainty? The view from classical decision theory is quite unequivocal:

*"First, the uncertainties present in the situation must be quantified in terms of values called probabilities. Second, the various consequences of the courses of action must be similarly described in terms of utilities. Third that decision must be taken which is expected - on the basis of the calculated probabilities - to give the greatest utility."*

<div align="right">Dennis Lindley [3].</div>

Probabilistic inference, or "how degrees of belief are altered by data"[4] is one of the two pillars of classical decision theory. Unfortunately it is widely acknowledged that objective probabilities (e.g. frequency based estimates of the cooccurrence of symptoms and diseases) are impractical for general decision making. Therefore the Bayesian notion of "subjective probability", of a person's willingness to accept a wager, has been formulated in a well-defined way in order to finesse this difficulty.

In reality the (psychological) processes that are involved in the formulation of subjective probabilities, and the formal nature of such numbers, are obscure. Probabilistic inference certainly places clear mathematical requirements on "coherent" belief revision procedures but it pays little attention to the question of what numerical degrees of belief can be said to represent. Unfortunately, heuristic methods may suffer from the reverse problem; while symbolic representational techniques are claimed to capture knowledge of informal domains like medicine quite well, the formal requirements of sound reasoning are not always adequately addressed.

The second pillar of decision theory is utility: roughly a numerical representation of the costs and benefits associated with deciding on a particular option. Unfortunately, as with probability, though for different reasons, it is often difficult to assign objective measures of utility to the consequences of decisions (e.g. the utility of life and death; pain or distress). Even in situations where there seems to be an objective scale (e.g. monetary value) the relationship between subjective and objective scales is not at all clear.

---

1. This figure is not untypical; in gastroenterology a diagnostic accuracy from the patient history alone is frequently much lower than this because of high intrinsic uncertainty.

It should also be noted that subjective values are multidimensional not unidimensional, and often qualitative. For example a drug treatment may be desirable because it is painless, because it can be taken at home, and because it is low cost, while it may be less desirable than a surgical procedure because the latter has a higher success rate, though it may compromise long term quality of life. Quantitative representation of values remains, at the very least, controversial.

In this section we will describe techniques for a number of aspects of decision making which are non-quantitative and yet can be clearly formulated. These techniques make use of first-order logic (FOL) to formulate methods for reasoning *for* and *against* decision options; introducing new options; structuring the decision; representing beliefs, values and preferences; taking the decision, and improving communication between decision support systems and their users.

## 2.1 EXTENDING THE REQUIREMENTS FOR SYMBOLIC REASONING

Inference is the pivot of most kinds of problem solving and decision making is no exception. Classical decision theory emphasises probabilistic inference, and many other techniques (both numerical and logical) are being developed to capture aspects of commonsense inference which are not expressible in standard monotonic logic [22]. However, these formalisations apply to just one (albeit a central one) of the activities associated with decision support. They address the evaluation and selection of decision options. Surrounding this activity are a number of further layers of activities, represented in figure 1 (after Andriole [5]), involving information acquisition, the actual identification of relevant decision options, and so forth. The concentration on the development of formalisms for option evaluation and selection has distracted attention from providing a more formal basis for these other activities [6].

To address the requirements raised by a more eclectic view of decision support we shall have to extend radically our notion of inference to one which can work at the multiplicity of levels represented in figure 1. We need to construct a general inference mechanism which satisfies at least the following requirements:

1. It must be able to construct arguments for decision options using whatever knowledge is productive; we do not wish to restrict reasoning to, nor for it to wholly depend upon, any one kind of inference (such as statistical inference) if this is restrictive.

2. There must be an explicit conceptualisation of *what it means to make a decision* and the roles of different kinds of inference in that process [7].

3. It is desirable to have a simple declarative representation of the decision procedure, decision criteria and application knowledge, permitting greater flexibility and more natural understanding by designers and users.

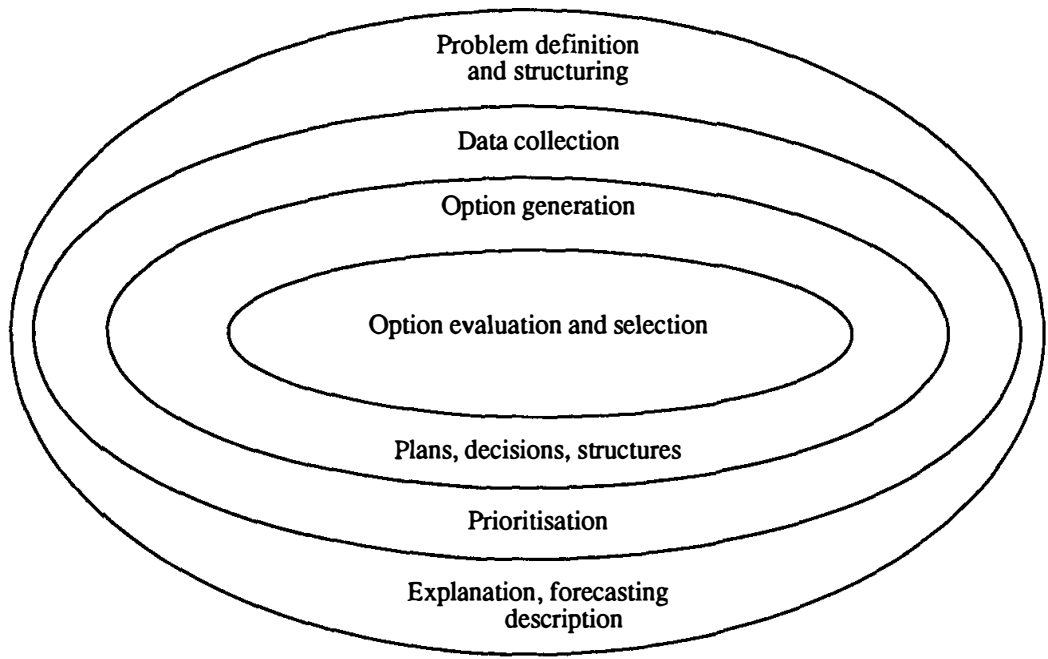

Figure 1: Decision making activities (after Andriole [5]).

4. Of course we also want our extended inference to be well understood and mathematically sound.

The demand for greater capabilities in decision support systems, and the need to address these requirements is encapsulated in a central goal of our research: achieving a deep understanding of what we have dubbed *symbolic decision procedures (SDPs)*. Informally a symbolic decision procedure can be characterised as:

*an explicit representation of the knowledge required to define, organise and make a decision, and ... a logical abstraction from the qualitative and quantitative knowledge that is required for any specific application. A SDP may include a specification of when and how the procedure is to be executed [6].*

The basic idea of classical decision making is the weighing of quantitative evidence and values for predefined options; the fundamental mechanism of a symbolic decision procedure is a flexible framework for qualitative reasoning that we call argumentation. Argumentation provides a rigorous basis for weighing the pros and cons of decisions but also for structuring the decision and, as we shall see later, for controlling the initiation and execution of a decision process.

## 2.2 INFERENCE AND "ARGUMENTATION"

Our central thesis is that practical decision making requires diverse sources of domain specific and domain independent knowledge. The relevance and nature of various forms of knowledge may vary with the context and the decision. To accommodate this, we use an extended form of inference, which we call argumentation. Informally we want to capture some of the kinds of argument that are commonly used in decision making, such as "based on the facts that the patient is *elderly* and suffering from *weight loss*, there is *support* for the proposition that *cancer is present*"

Argumentation consists of the identification and appraisal of lines of reasoning about propositions. Argumentation permits the use of whatever theories and sources of knowledge are deemed appropriate. An argument for a proposition P may have an associated *qualifier* or *sign* $S_p$ which represents the certainty of the proposition given that the argument is valid. The *grounds* for the argument, $G_p$, are also associated with the proposition. These indicate the *facts* used in initiating the argument, and the *theories* used to connect these facts with P.

More formally, if an argument is identified for a proposition P, then we may write

$$KB \vdash (P, S_p, G_p)$$

with $S_p$ and $G_p$ as above. The qualifier $S_p$ and grounds $G_p$ are not merely present to provide useful information to the user. They also satisfy a more formal requirement. Multiple triples will be associated with a proposition P if P can be deduced with different qualifiers $S_p$, or with the same qualifiers but using different grounds $G_p$. We say these constitute distinct arguments for P.

The knowledge base, KB, over which the inference engine operates may be partitioned into a number of theories. These may be domain specific theories, consisting primarily of ground facts, or domain independent theories consisting of first order rules for establishing links between knowledge items in a given task. A further knowledge layer may contain information about which theories are relevant to which specific tasks. In our view practical decision making may involve arguing from different points of view (e.g. arguing from a causal theory, an anatomical theory, a theory of physiological function or from statistical knowledge). The kinds of theory that are relevant are determined by the kind of decision that is being taken. We have to keep these theories partitioned because in practice we cannot guarantee that the different views that they embody are globally consistent.

A more detailed discussion on how arguments can be constructed may be found in [9] and [26].

## 2.3 AGGREGATION OF ARGUMENTS

We have seen that from different grounds it is possible to construct multiple arguments about (for or against) a proposition. If:

$$KB \vdash (P, S_1, G_1)$$
$$KB \vdash (P, S_2, G_2)$$

where $S_1 \neq S_2$, or $G_1 \neq G_2$, then there are two distinct arguments for P. Once two or more distinct arguments have been identified for a proposition P, the associated qualifiers $S_i$ will need to be aggregated to form a global qualification of P.

The process of argumentation allows plenty of scope for the definition of various aggregation operators, both numerical and qualitative. The $S_i$ may be taken from a variety of symbolic or numeric dictionaries. Here we shall focus on a simple 4-valued dictionary:

{confirmed, eliminated, supported, opposed}.

exactly one of which must be associated with each argument. The semantics of these terms is not numerical, but where probabilities (or other numerical certainty data) are available they could be substituted without modification to the basic argumentation framework.

Our interest of course is in the more common situation where those data are not available. A straightforward aggregation operator is simply to add up the arguments for and against each option, and decide on that with the largest ratio of pros to cons. This was the method employed in the gastroenterology example described in the introduction, and such improper linear decision rules are well recognised as effective.

We can also formulate symbolic aggregation operators. Making the grounds explicit in the argument has obvious value in the user interface, but they  can also play an im-portant part in aggregation because we can use them in computing states of belief in options (or any proposition). This can be done using logical schemata which demand no numerical coefficients, but which we believe are intuitively appealing and logically coherent.

The following symbolic aggregation rules[1] define logical schemata for assigning propositions to various classes of belief: *conceivable*; *possible*; *plausible*; *confirmed*.

(P, conceivable)
  ← not ∃G • (P, G, eliminated).

(P, possible)
  ← (P, conceivable) ∧ ∃G • (P,G, supported).

(P, plausible)
  ← (P, possible) ∧ not ∃G • (P, G, opposed).

(P, confirmed)
  ← (P, conceivable) ∧ ∃G • (P, G, confirmed).

The aggregation operator defined by this schema allows only a fairly coarse-grained categorisation of decision options. A more complex schema could be defined to allow finer distinctions, but in *The Oxford System of Medicine* (OSM), a decision support system designed to provide flexible assistance for general medical practice (and a specialised derivative for oncology, BOSS) [16], the 4 qualifiers in the dictionary above, together with the belief terms constructible from them, provide an adequate basis for carrying out decision making.

Our contention is that for many problems, where limited statistical data are available, attempts to use numbers to make a fine grained distinction between decision options may be unnecessary, and lead to illusory precision in the final result. A logical approach to arguing for and against and comparing decision options, on the other hand, has a clearly defined semantics which is easily explained to and understood by the user.

### 2.4 DISCUSSION

Current probabilistic techniques require the prior construction of the graph linking decision options to observables and findings[2]. However, the exact nature of these links may vary with context and the nature of the decision problem at hand, and in general decision systems should be able to construct and revise the graph dynamically as information becomes available and the goals of the problem at hand are identified [15]. Our approach allows for arguing about this graph structure [14],[15]; for example, the discovery that a patient is under medication which can cause an observed abnormality as a side effect may lead to a revision of the graph that had previously been generated.

We have described a purely symbolic approach to identifying and evaluating decision options. Further to this, we aim to develop techniques which address how and when a decision may be taken. As with symbolic terms for representing states of belief, symbolic representations of preference and value may not be arbitrary but must be assigned an explicit semantics. In order to avoid confounding distinct logical ideas like obligations, duties and preferences we are taking an approach in which arguments are constructed from principles of what *must* be done ("deontic" principles) and what *ought* to be done ("praxeological" principles) and pluralistic value theory [10].

It is our intention to strive towards a normative theory of decision making. To achieve this we need "an explicit conceptualisation of what it means to make a decision" (requirement 2 above). A possible approach to this is discussed in the next section.

## 3  AUTONOMOUS DECISION MAKING UNDER UNCERTAINTY

The central claim of section 2 was that symbolic procedures can significantly extend the capabilities for reasoning about belief and values and structuring the decision as compared with strictly numerical procedures (requirement 1 above). A further advantage we claim is that a symbolic approach allows for an explicit representation of the decision itself. Neither classical decision theory nor work on knowledge based decision aids have placed emphasis on this, apparently because these systems are intended to *assist* in decision making, not to make decisions autonomously. They can rely on knowledgeable users to critically supervise the decision process. In our view this is a serious omission from, and challenge to, theory. For practical reasons and to be confident in our understanding of the limits on decision making capabilities we need to address the problem of building systems that can operate autonomously, without relying on external support.

There is a steadily increasing interest in AI in the development of autonomous agents [23]. In particular SOAR, an "architecture for general intelligence" [21] and HOMER, a simulated submersible capable of receiving task instructions and autonomously planning its solution [24], are projects which are making interesting progress towards autonomous capabilities. Although not developed with either decision theory or uncertainty management in mind they may guide us towards a statement of what the capabilities of an autonomous decision system should be. We first attempt the following definition of an autonomous agent:

*An agent is autonomous with respect to its environment if it can set and achieve goals, and respond adaptively to events in its environment, without external advice or assistance.*

Practical environments are frequently so complex they can evolve in far more ways than could be allowed for in any *a priori* structuring of a decision. A definite requirement for an autonomous decision maker therefore is that it should be responsive to the arrival of unexpected information. For example, HOMER [24] is given an instruction to collect a package from a pier and constructs a plan to do this. *En route*, however, it finds a large ship on its course; HOMER must perceive this, recognise its implications and replan to achieve its goal.

---
1. Called "annotation rules" in [6], [14], [15].
2. Although we are grateful to an anonymous referee for drawing our attention to work that is underway to correct this deficiency.

In general information may at any time become available to a decision maker that has implications for any aspect of a decision *viz*: raising new problems requiring additional decisions; challenging the grounds for current beliefs, or indicating that the current decision can be taken without further information.

Part of the responsiveness of an effective decision maker is the ability to recognise that there is a decision to be taken. The *Concise Oxford Dictionary* relates a decision to the "settlement of a question". Classical decision theory has had much to say about how we should settle a question; the new challenge is to understand how these questions are formulated in the first place. This suggests a revision to our definition:

*An agent is an autonomous decision maker if it can pose and resolve questions about the state of its environment or the actions that are desirable to achieve its goals*

If a decision maker could ask and answer questions such as the following it would gain great power:

1. The pivotal questions are: *what is the problem? what do I need to know, or do?* As with all AI systems we will necessarily require an explicit representation of the systems goals in order to be able to formulate these questions. Decision theory has taken the question of what a decision is required for entirely for granted. As soon as we address it a cascade of further questions follows.

2. *What do I know that is relevant to this decision?* An agent that has a great deal of knowledge may encounter difficulties in retrieving relevant knowledge during decision making. Explicit representation of theories and their applicability aids search.

3. *What are the possible options? How could I find out?* Symbolic decision procedures can introduce decision options as information is acquired by making use of explicit advice like "in diagnosis, propose possible causes of symptoms as possible diagnoses".

4. *What justifies a belief (value or preference)? Is this argument still valid?* Practical problem solving almost always risks blind-alleys and misunderstandings. The underpinning of beliefs by explicit arguments that record the grounds for those beliefs provides the information necessary for detecting and resolving inconsistencies [25].

5. *Am I thinking about this decision the right way?* All problem solving takes place in a "problem space" which embodies presuppositions about the problem and affects the way problem solving proceeds. If presuppositions or representations are implicit then we are at the mercy of them. If theories and presuppositions are explicit then the agent potentially has the ability to detect when assumptions of validity or relevance are violated.

### 3.1 REQUIREMENTS AND SKETCH OF AN AUTONOMOUS DECISION MAKER

In this section we summarise some of the principle requirements for a comprehensive decision capability and the main theoretical challenges that they entail. The principle requirements for an autonomous decision maker include the following:

1. It should be able to observe and interpret its environment, and recognise when a decision or sequence of decisions needs to be taken in order to achieve its goals.

2. Goals and decisions should be represented explicitly. One approach (used in the OSM and BOSS) is to represent the "generic" decision as the root class in a generalisation hierarchy. This defines the decision procedure as a set of partially instantiated attribute templates, such as:

    decision_prototype(Decision,Prototype).
    relevant_theories(Decision,Theories).
    relevant_argument_types(Decision,Arguments).
    option_proposal_criteria(Decision,Criteria). ...

3. The decision maker should be able to classify the *types* of decision required. One way to do this is to associate with each class a particular decision_prototype, invoking it when its prototype is satisfied. For example a prototype for a diagnosis decision may specify that if an observation has been made that is abnormal and its cause is not known then a decision of class "diagnosis" is required in the context. Specific classes of decision inherit the generic attributes, but are distinguished from it by the values that instantiate the attribute templates, as in:

    relevant_theories(diagnosis, symptomatology).
    relevant_argument_types(diagnosis,causality).
    option_proposal_criteria(diagnosis,possibility) ...

4. The agent must be able to initiate the decision in the context; this will entail inheriting all the class information to the decision instance, further instantiating it with details of the context (e.g. a patient's name) and presumably executing some control actions.

5. Guiding the decision process. Explicit knowledge of relevant theories and arguments associated with the decision class can be used to guide information acquisition and data interpretation. Decision options (e.g. diagnoses) can be proposed on the basis of criteria associated with the class (in the above example any option is proposed as a candidate if it satisfies the condition for being "possible" (section 2).

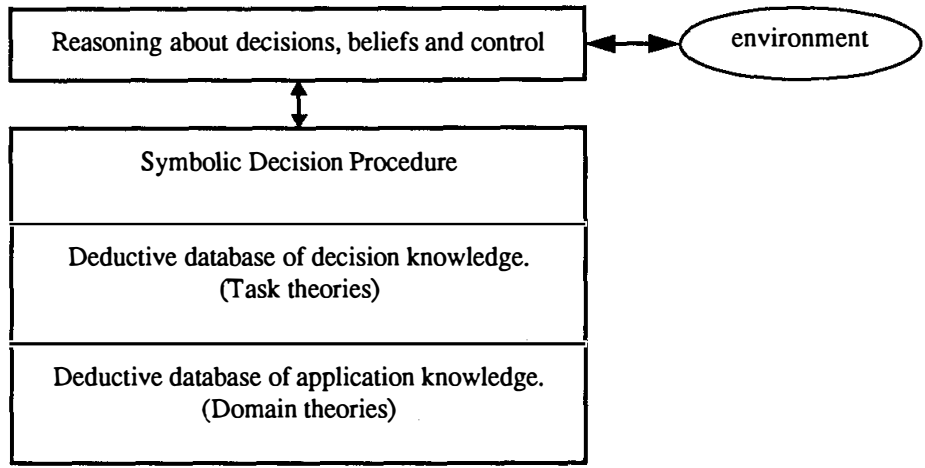

Figure 2: Outline architecture for autonomous decision agents.

6. The decision should be terminated when appropriate. Explicit criteria for formulating decisions also presumably imply criteria for knowing when those decisions can be taken. For example, the criterion given above for requiring a diagnosis decision was that we cannot confirm the cause of an abnormal observation. The opposite is also true; if we can confirm the cause then the diagnosis can be made.

Work is in progress to provide an adequate theoretical basis for these capabilities, and to demonstrate their application. Figure 2 presents a schematic outline of a proposed layered architecture for autonomous decision agents. The ascending layers represent knowledge at increasing levels of abstraction; each layer is capable of manipulating the knowledge in the layers beneath it.

The three lower layers are deductive data bases representing (in ascending order) domain theories (application specific knowledge), task theories (knowledge about decisions), and the symbolic decision procedure itself. The top layer in figure 2 may be described as a control layer; among its functions are to look after interactions with the environment, respond to events, initiate, control and terminate decisions and arguments, and maintain the agent's beliefs and goals. This organisation echoes the traditional distinction between knowledge and action; the symbolic decision procedure and other deductive components can be naturally implemented with a pure logic theorem prover, while the top layer - which operates in time, entails side-effects on the system's knowledge, and has other non-logical features - is procedural in character.

A full explanation of these examples would need a detailed description of an interpreter, which cannot be presented here. Some discussion of techniques to provide this capability can be found in [27].

## 4  PRINCIPLE RESEARCH DIRECTIONS

It is important to carry out work like this in the context of applications. The OSM and BOSS systems, under development in this laboratory, have been important in forcing us to develop our approaches and in evaluating our ideas. For generality we would like to see applications in domains other than medicine and for decision tasks other than medical decisions. We also hope to build on our approach to look into *compound decisions*, such as planning, design and other tasks. These tasks can be viewed (in part) as complexes of simple selection decisions. Some experimental work on integrating decisions in problem solving is in progress, focusing on planning of cancer treatment and formulation of antibiotic therapy for chest infections [19].

We are in the early stages of formalising argumentation and other aspects of symbolic decision making, focusing on non-autonomous decision support systems. Some work on autonomous systems is in progress, and influences the work, but for the moment is of lower priority.

Perhaps one of the most obvious features of our proposals is the importance we attach to meta-level reasoning, or reflection. To express the questions in section 3 and implement the mechanisms required in section 3.1 it seems clear that considerable capabilities for reflection are needed - reflecting on beliefs, arguments, knowledge, and decisions. Classical decision procedures offer few handholds for developing these ideas. Formalising the concept of reflection will be, we believe, fundamental to significant progress in the field. It is perhaps the most difficult yet most fascinating theoretical challenge before us.

110  Fox and Krause

Acknowledgements

We would like to thank all our colleagues in the Biomedical Computing Unit for many useful discussions on this work, but especially Saki Hajnal, Dominic Clark, Andrzej Glowinski, Mike O'Neil and Mirko Dohnal. This work has received an added stimulus with our involvement in and support from the Esprit Basic Research Programme 3085, DRUMS.

P. Krause is supported under the SERC project 1822: a Formal Basis for Decision Support Systems. Our thanks also to Mike Clarke of Queen Mary College who is involved in this project with us.